\documentclass[10pt,twocolumn,letterpaper]{article}

\usepackage{cvpr}
\usepackage{times}
\usepackage{graphicx}
\usepackage{amsmath}
\usepackage{amssymb}
\usepackage{booktabs}
\usepackage{multirow,array}
\usepackage[usenames,dvipsnames]{color}
\usepackage{bm}
\usepackage{subcaption}

 \cvprfinalcopy % *** Uncomment this line for the final submission

\def\cvprPaperID{862} % *** Enter the CVPR Paper ID here

\ifcvprfinal\fi

\newcommand{\C}{\mathcal{C}}
\newcommand{\weak}{\mathcal{W}}
\newcommand{\strong}{\mathcal{S}}
\newcommand{\R}{R}
\newcommand{\bagL}{Y}
\newcommand{\instL}{y}
\newcommand{\feat}{\bm{x}}
\newcommand{\w}{\bm{w}}

\begin{document}

%%%%%%%%% TITLE
%\title{Using All  the Data: \\Joint Optimization of Detectors with Weakly and Strongly Annotated Data}
%\title{Why Not Re-use Annotations? Weak Detector Learning \\Regularized by Strong Annotations from Auxiliary Tasks}
%\title{Why Ignore Annotations? Weak Detector Learning \\Regularized by Strong Annotations from Auxiliary Tasks}
%\title{Large Scale Detector Learning: A Joint Optimization \\Framework Using Strong and Weak Annotations}
\title{Detector Discovery in the Wild: \\Joint Multiple Instance and Representation Learning}
% Strongly or Weakly Supervised Learning, why not both?
% Using all the data
% Let's 

\author{Judy Hoffman, Deepak Pathak, Trevor Darrell\\
UC Berkeley\\
{\tt\small \{jhoffman, pathak, trevor\}@eecs.berkeley.edu}
\and Kate Saenko\\
UMass Lowell\\
{\tt\small saenko@cs.uml.edu}
}

\maketitle
%\thispagestyle{empty}

%%%%%%%%% ABSTRACT
\begin{abstract}
We develop methods for detector learning which exploit joint training over both weak and strong labels and which transfer learned perceptual representations from strongly-labeled auxiliary tasks. Previous methods for weak-label learning often learn detector models independently using latent variable optimization, but fail to share deep representation knowledge across classes and usually require strong initialization. Other previous methods transfer deep representations from domains with strong labels to those with only weak labels, but do not optimize over individual latent boxes, and thus may miss specific salient structures for a particular category.  We propose a model that subsumes these previous approaches, and simultaneously trains a representation and detectors for categories with either weak or strong labels present. We provide a novel formulation of a joint multiple instance learning method that includes examples from classification-style data when available, and also performs domain transfer learning to improve the underlying detector representation. Our model outperforms known methods on ImageNet-200 detection with weak labels.
\end{abstract}

%%%%%%%%% BODY TEXT
\section{Introduction}
\label{sec:intro}
% Introduction

\begin{figure}[t]
\begin{center}
   \includegraphics[width=\linewidth, natwidth=1400, natheight=910]{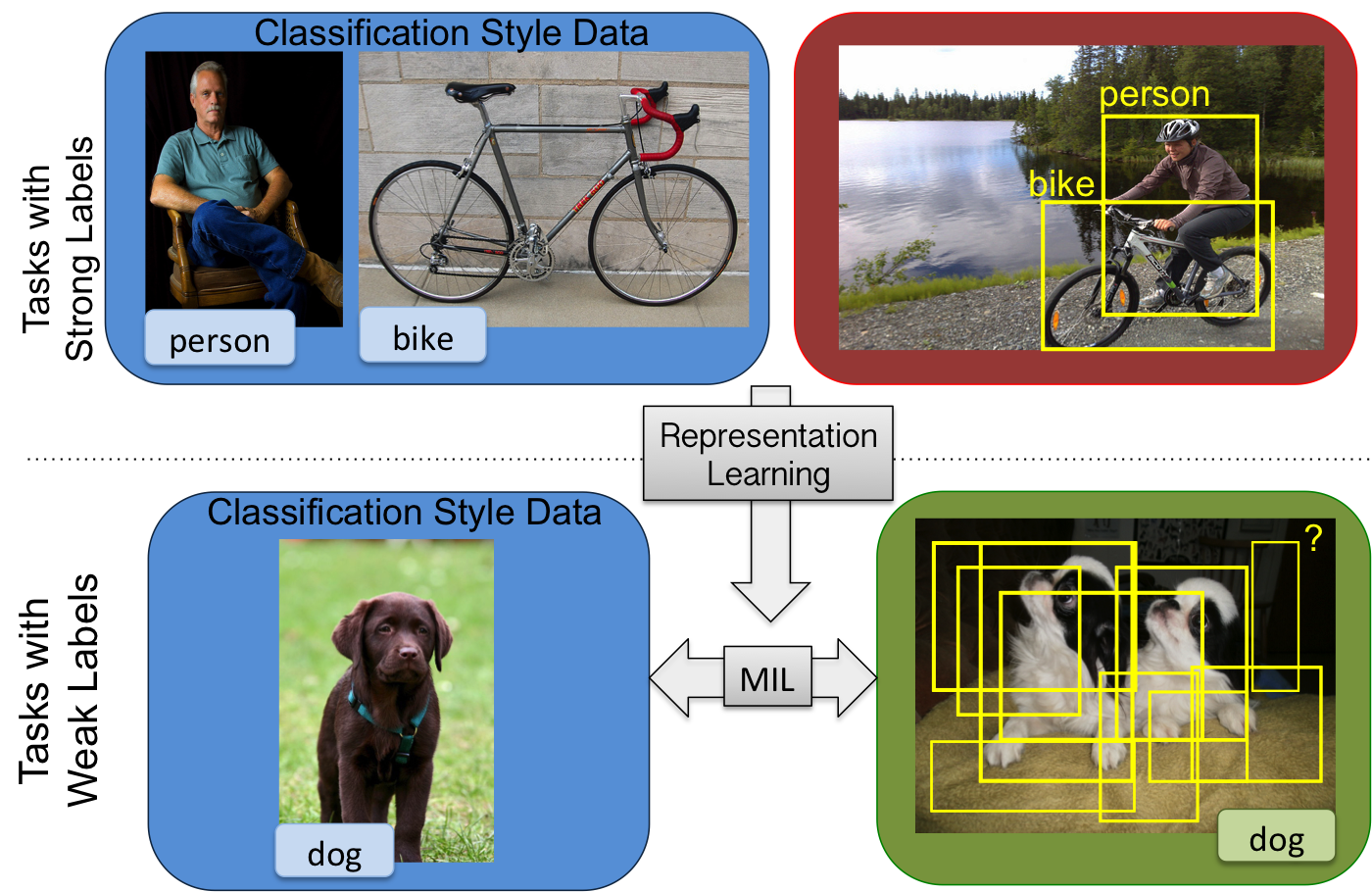}
\end{center}
\caption{
We learn detectors for categories with only weak labels (\emph{bottom row}), by jointly transferring a representation from auxiliary categories with available strong annotations (\emph{top row}) and solving an MIL problem on the weakly annotated data (green box). 
}
\label{fig:overview}
\end{figure}

It is well known that contemporary visual models thrive on large amounts of training data, especially those that directly include labels for desired tasks.  Many real world settings contain labels with varying specificity, e.g., ``strong" bounding box detection labels, and ``weak" labels indicating presence somewhere in the image. We tackle the problem of \emph{joint detector and representation learning}, and develop models which cooperatively exploit heterogeneous sources of training data, where some classes have no ``strong" annotations.  Our model optimizes a latent variable multiple instance learning model over image regions while simultaneously transferring a shared representation from detection-domain models to classification-domain models. The latter provides a key source of automatic and accurate initialization for latent variable optimization, which has heretofore been unavailable in such methods.

Previous methods employ varying combinations of weak and strong labels of the same object category to learn a detector. Such methods seldom exploit available strong-labeled data of different, auxiliary categories, despite the fact that such data is very often available in many practical scenarios. Deselaers \etal~\cite{deselaers2012weakly} uses auxiliary data to learn generic objectness information just as an initial step, but doesn't optimize jointly for weakly labeled data. 

We introduce a new model for large-scale learning of detectors that can jointly exploit weak and strong labels, perform inference over latent regions in weakly labeled training examples, and can transfer representations learned from related tasks (see Figure~\ref{fig:overview}).  In practical settings, such as learning visual detector models for all available ImageNet categories, or for learning detector versions of other defined categories such as Sentibank's adjective-noun-phrase models \cite{sentibank}, our model makes greater use of available data and labels than previous approaches.  Our method takes advantage of such data by using the auxiliary strong labels to improve the feature representation for detection tasks, and uses the improved representation to learn a stronger detector from weak labels in a deep architecture.

To learn detectors, we exploit weakly labeled data for a concept, including both ``easy" images (e.g., from ImageNet classification training data), and ``hard" weakly labeled imagery (e.g., from PASCAL or ImageNet detection training data with bounding box metadata removed).  We define a novel multiple instance learning (MIL) framework that includes bags defined on both types of data, and also jointly optimizes an underlying perceptual representation using strong detection labels from related categories.  The latter takes advantage of the empirical results in \cite{lsda}, which demonstrated knowledge of what makes a good perceptual representation for detection tasks could be learned from a set of paired weak and strong labeled examples, and the resulting adaptation could be transferred to new categories, even those for which no strong labels were available.

We evaluate our model empirically on the largest set of available ground-truth visual detection data, the ImageNet-200 category challenge.  Our method outperforms the previous best MIL-based approaches for held-out detector learning on ImageNet-200~\cite{ilsvrc} by 200\%, and outperforms the previous best domain-adaptation based approach \cite{lsda} by 12\%. Our model is directly applicable to learning improved ``detectors in the wild", including categories in ImageNet but not in ImageNet-200, or categories defined ad-hoc for a particular user or task with just a few training examples to fine-tune a new classification model.  Such models can be promoted to detectors with no (or few) labeled bounding boxes.  Upon acceptance we will release an open-source implementation of our model and all network and detector weights for an improved set of detectors for the ImageNet-7.5K dataset of \cite{lsda}.

\section{Related Work}
\label{sec:related}
% Related work
\paragraph{CNNs for Visual Recognition} Within the last few years, convolutional neural networks (CNNs) have emerged as the clear winners for many visual recognition tasks. A breakthrough was made when the positive performance demonstrated for digit recognition~\cite{lecun89} began to translate to the ImageNet~\cite{ilsvrc} classification challenge winner~\cite{supervision}. Shortly thereafter, the feature space learned through these architectures was shown to be generic and effective for a large variety of visual recognition tasks~\cite{decaf, zeiler-arxiv-2013}. These results were followed by state-of-the-art results for object detection~\cite{rcnn, overfeat}. Most recently, it was shown that CNN architectures can be used to transfer generic information between the classification and detection tasks~\cite{lsda}, improving detection performance for tasks which lack bounding box training data.

\paragraph{Training with Auxiliary Data Sources}
There has been a large amount of prior work on training models using auxiliary data sources. The problem of visual domain adaptation is precisely seeking to use data from a large auxiliary source domain to improve recognition performance on a target domain which has little or no labeled data available. Techniques to solve this problem consist of learning a new feature representation that minimizes the distance between source and target distributions~\cite{saenko-eccv10, kulis-cvpr11, gong-cvpr12, sa}, regularizing the learning of a target classifier against the source model~\cite{asvm, pmt, daume}, or doing both simultaneously~\cite{hoffman-iclr13, duan-icml12}.

%%%%%% Overall Method Fig -- Placing here to look nicer %%%%%%%
\begin{figure*}
\centering
\includegraphics[width=\linewidth, natwidth=1685, natheight=512]{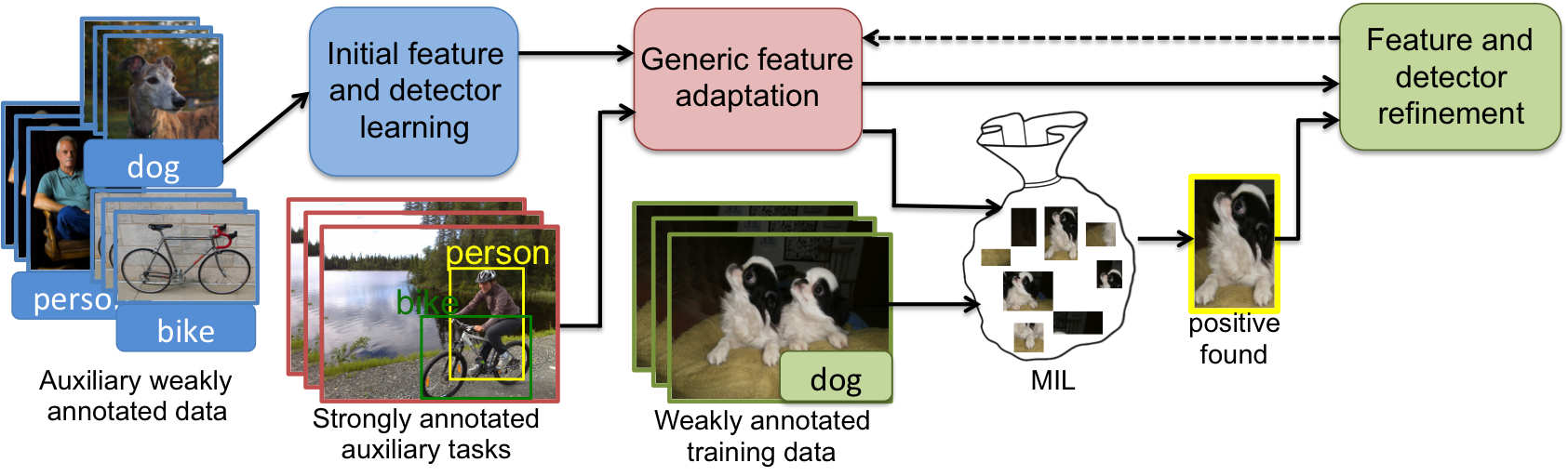}
\caption{Our method jointly optimizes a representation and detectors for categories with only weakly annotated data. 
We first learn a feature representation conducive to MIL by initializing all parameters with classification style data. 
%and then refining the feature space with strongly annotated data from auxiliary tasks. We finally perform MIL in our detection feature space and use the discovered positive patches to further refine the representation and detection weights.
We then collectively refine the feature space with strongly annotated data from auxiliary tasks, and perform MIL in our detection feature space. The discovered positive patches are further used to refine the representation and detection weights.
}
\label{fig:method}
\end{figure*}
%%%%%%%%%%%%%%%%%%%%%%%%%%%%%%%%%%%%%%%

\paragraph{Multiple Instance Learning}
Since its inception, the MIL~\cite{dietterich1997solving} problem has been attempted in several frameworks including Noisy-OR~\cite{heckerman2013tractable}, boosting~\cite{ali_cvpr14,zhang2005multiple} etc. But most commonly, it was framed as a max-margin classification problem~\cite{andrews2002_nips} with latent parameters optimized using alternating optimization~\cite{felzenszwalb2010_pami,thorsten2009_icml}. 
Overall, MIL is tackled in two stages: first finding better initialization, and then using better heuristics for optimization. 
A number of methods have been proposed for initialization which include using large image region excluding boundary~\cite{pandey2011scene}, using candidate set which covers the training data space~\cite{song2014_icml,song2014weakly}, using unsupervised patch discovery~\cite{siva2013looking,singh2012unsupervised}, learning generic objectness knowledge from auxiliary catgories~\cite{alexe2010object,deselaers2012weakly}, learning latent categories from background to suppress it~\cite{wang_eccv14} or using class-specific similarity~\cite{siva2012defence}. Approaches to better optimize the non-convex problem involve using multi-fold learning as a measure of regularizing overfitting~\cite{verbeek2014_cvpr}, optimize Latent SVM for the area under the ROC curve (AUC)~\cite{bilen2014_ijcv} and training with easy examples in beginning to avoid bad local optimization~\cite{bengio2009curriculum,kumar2010self}. Most of these approaches perform reasonably only when object covers most of the region of image, or when most of the candidate regions contain an object. The major challenge faced by MIL in general is that of fixed feature representation, and poor initialization particularly in non-object centric images.
Our algorithm provides solutions to both of these issues.

%This approach have also been formulated as inference of discriminant function over the feature representation which, in addition to input image, also depends on a latent parameter~\cite{felzenszwalb2010_pami,thorsten2009_icml}. The latent parameter corresponds to the location of object in the image. This is termed as Latent SVM, and is equivalent in principle to MI-SVM. Yu and Joachims~\cite{thorsten2009_icml} presents a general framework for latent parameter estimation in Structural SVMs. It bounds the loss function such that the final optimization equation can be represented as difference of convex functions, allowing to use classic concave-convex procedure (CCCP)~\cite{yuille2003_nc} which is similar to the alternating optimization heuristic described above. In this paper, we use the Latent SVM formulation~\cite{thorsten2009_icml} and optimize it for area under the ROC curve (AUC) as suggested in Bilen \etal~\cite{bilen2014_ijcv}.  Since there are lot more negative examples in general, AUC optimization is robust to data imbalance and also improves the performance on testing criteria i.e. average precision.

\section{Background: MI-SVM}
\label{sec:method-mil}
% Method -MIL background
We begin by briefly reviewing a standard solution to the multiple instance learning problem, Multiple Instance SVMs (MI-SVMs)~\cite{andrews2002_nips} or Latent SVMs~\cite{felzenszwalb2010_pami,thorsten2009_icml}. 
In this setting, each weakly labeled image is considered a collection of regions which form a positive `bag'.  
For a binary classification problem, the task is to maximize the bag margin which is defined by the instance with highest confidence. 
For each weakly labeled image $I\in \weak$, we collect a set of regions of interest and define the index set of those regions as $\R_I$. We next define a bag as $B_I = \{\feat_i | i\in \R_I\}$, with label $\bagL_I$, and let the $i^{th}$ instance in the bag be $(\feat_i,\instL_i)\in \mathcal{R}^p\times \{-1,+1\}$.

For an image with a negative image-level label, $\bagL_I = -1$, we label all regions in the image as negative. For an image with a positive image-level label, $\bagL_I = 1$, we create a constraint that at least one positive instance occurs in the image bag. 

In a typical detection scenario, $\R_I$ corresponds to the set of possible bounding boxes inside the image, and maximizing over $\R_I$ is equivalent to discovering the bounding box that contains the positive object. We define a representation $\phi(\feat_i)\in \mathcal{R}^d$ for each instance, which is the feature descriptor for the corresponding bounding box, and formulate the MI-SVM objective as follows:
\begin{equation}\label{eq:misvm}
\min_{\w \in \mathcal{R}^d} \quad \frac{1}{2} \|\w \|_2^2 + \alpha \sum_{I}\ell\Big(\bagL_I , \max_{i\in \R_I} \w^T \phi(\feat_i) \Big) 
\end{equation}\
where $\alpha$ is a hyper-parameter and $\ell(\instL,\hat{\instL})$ is the hinge loss.
Interestingly, for negative bags i.e. $\bagL_I=-1$, the knowledge that all instances are negative allows us to unfold the max operation into a sum over each instance. Thus, Equation~\eqref{eq:misvm} reduces to a standard QP with respect to $\w$. For the case of positive bags, this formulation reduces to a standard SVM if maximum scoring instance is known. 

Based on this idea, Equation~\eqref{eq:misvm} is optimized using a classic concave-convex procedure~\cite{yuille2003_nc}, 
which decreases the objective value monotonically with a guarantee to converge to a local minima or saddle point. 
Due to this reason, these methods are extremely susceptible to the feature representation and detector initialization~\cite{verbeek2014_cvpr,song2014_icml}. 
We address both these issues using annotated auxiliary data available to learn a better feature representation and reasonable initialization for MIL based methods.

\section{Large Scale Detection Learning}
\label{sec:method}
% Method

We propose a detection learning algorithm that uses a heterogeneous data source, containing only weak labels for some tasks, to produce strong detectors for all.  
Let the set of images with only weak labels be denoted as $\weak$ and the set of images with strong labels (bounding box annotations) from auxiliary tasks be denoted as $\strong$. 
We assume that the set of object categories that appear in the weakly labeled set, $\C_{\weak}$, do not overlap with the set of object categories that appear in the strongly labeled set, $\C_{\strong}$. 
For each image in the weakly labeled set, $I\in \weak$, we have an image-level label per category, $k$: $\bagL_I^k \in \{1,-1\}$. For each image in the strongly labeled set, $I \in \strong$, we have a label per category, $k$, per region in the image, $i \in \R_I$: $\instL_{i}^k \in \{1,-1\}$. 
We seek to learn a representation, $\phi(\cdot)$ that can be used to train detectors for all object categories, $\C = \{\C_{\weak} \cup \C_{\strong}\}$. For a category $k \in \C$, we denote the category specific detection parameter as $\w_k$ and compute our final detection scores per region, $\feat$, as $score_k(\feat) = \w_k^T \phi(\feat)$.

% Joint objective 
We propose a joint optimization algorithm which learns a feature representation, $\phi(\cdot)$, and detectors, $w_k$, using the combination of strongly labeled detection data, $\strong$, with weakly labeled data, $\weak$. 
For a fixed representation, one can directly train detectors for all categories represented in the strongly labeled set, $k \in C_{\strong}$. Additionally, for the same fixed representation, we reviewed in the previous section techniques to train detectors for the categories in the weakly labeled data set, $k \in \C_{\weak}$. Our insight is that the knowledge from the strong label set can be used to help guide the optimization for the weak labeled set, and we can explicitly adapt our representation for the categories of interest and for the generic detection task. 

Below, we state our overall objective:
\begin{eqnarray}
\min_{\substack{\w_k,\phi\\k \in \C}} & & \sum_{k}\Gamma(\w_k) \label{eq:joint}\\
		&& + \alpha \sum_{I \in \weak} \sum_{p\in \C_{\weak}} \ell(\bagL^p_I, \max_{i\in \R_I} \w_p^T \phi(\feat_i)) \nonumber \\
		&& + \alpha \sum_{I \in \strong} \sum_{i \in \R_I} \sum_{q \in \C_{\strong}} \ell(\instL^q_i, \w_q^T\phi(\feat_i))\nonumber
\end{eqnarray}
where $\alpha$ is a scalar hyper-parameter, $\ell(.)$ is the loss function and $\Gamma(.)$ is a regularization over the detector weights. 
This formulation is non-convex in nature due to the presence of instance level ambiguity. It is difficult to optimize directly, so we choose a specific alternating minimization approach (see Figure~\ref{fig:method}).

We begin by initializing a feature representation and initial CNN classification weights using auxiliary weakly labeled data (blue boxes Figure~\ref{fig:method}). These weights can be used to compute scores per region proposal to produce initial detection scores. We next use available strongly annotated data from auxiliary tasks to transfer category invariant information about the detection problem. We accomplish this through further optimizing our feature representation and learning a generic background detection weights (red boxes Figure~\ref{fig:method}). We then use the well tuned detection feature space to perform MIL on our weakly labeled data  to find positive instances (yellow box Figure~\ref{fig:method}. Finally, we use our discovered positive instances together with the strongly annotated data from auxiliary tasks to jointly optimize all parameters corresponding to feature representation and detection weights.

\subsection{Initialize Feature Representation and Detector Weights}
We now discuss 
our procedure for initializing the feature representation and detection weights. We want to use a representation which makes it possible to separate objects of interest from background and makes it easy to distinguish different object categories. Convolutional neural networks (CNNs) have proved effective at providing the desired semantically discriminative feature representation~\cite{decaf, rcnn, overfeat}. We use the architecture which won the ILSVRC2012 classification challenge~\cite{supervision}, since it is one  of the best performing and most studied models.  The network contains roughly 60 million parameters, and so must be pre-trained on a large labeled corpus. Following the standard protocol, we use auxiliary weakly labeled data that was collected for training a classification task for this initial training of the network parameters (Figure~\ref{fig:method}: \emph{blue boxes}). This data is usually object centric and is therefore effective for training a network that is able to discriminate between different categories. We remove the classification layer of the network and use the output of the fully connected layer, $fc_7$, as our initial feature representation, $\phi(\cdot)$.

We next learn initial values for all of the detection parameters, $\w_k,\; \forall k\in \C$. To solve this, we begin by solving the simplified learning problem of image-level classification. 
The image, $I \in \strong$, is labeled as positive for a category $k$ if any of the regions in the image are labeled as positive for $k$ and is labeled as negative otherwise, we denote the image level label as in the weakly labeled case: $\bagL_I^k$. Now, we can optimize over all images to refine the representation and learn category specific parameters that can be used per region proposal to produce detection scores:
\begin{eqnarray}
\min_{\substack{\w_{k},\phi\\k \in \C} }  \sum_{k} \left[\Gamma(\w_k) 
		 + \alpha \sum_{I \in \{\weak\cup \strong\}} \ell(\bagL^k_I,\w_k^T \phi(I)) \right] \label{eq:classification}
\end{eqnarray}
We optimize Equation~\ref{eq:classification}  through fine-tuning our CNN architecture with a new $K$-way last fully connected layer, where $K=|\C|$.
 
\subsection{Optimize with Strong Labels From Auxiliary Tasks}

Motivated by the recent representation transfer result of Hoffman et al.~\cite{lsda} - LSDA, we learn to generically transform our classification feature representation into a detection representation by using the strongly labeled detection data to modify the representation, $\phi(\cdot)$, as well as the detectors, $\w_k, \; k \in \C_{\strong}$ (Figure~\ref{fig:method} : \emph{red boxes}). In addition, we use the strongly annotated detection data to initialize a new ``background" detector, $\w_b$. This detector explicitly attempts to recognize all data labeled as negative in our bags. However, since we initialize this detector with the strongly annotated data, we know precisely which regions correspond to background. The intermediate objective is:
\begin{eqnarray}
\min_{\substack{\w_q,\phi \\q \in \{\C_{\strong}, b\}}} \sum_{q} \Bigg[ \Gamma(\w_q) + \alpha \sum_{I \in \strong} \sum_{i \in \R_I} \ell(\instL^q_i, \w_q^T\phi(\feat_i)) \Bigg]\label{eq:strong}
\end{eqnarray}
Again, this is accomplished by fine-tuning our CNN architecture with the strongly labeled data, while keeping the detection weights for the categories with only weakly labeled data fixed. Note, we do not include the last layer adaptation part of LSDA, since it would not be easy to include in the joint optimization. Moreover, it is shown that the adaptation step does not contribute significantly to the accuracy~\cite{lsda}.
 
\subsection{Jointly Optimize using All Data}
 
With a representation that has now been directly tuned for detection, we fix the representation, $\phi(\cdot)$ and consider solving for the regions of interest in each weak labeled image.
This corresponds to solving the second term in Equation~\eqref{eq:joint}, i.e.:
\begin{eqnarray}
\min_{\substack{\w_p\\ p \in \{\C_{\weak}, b\}} } & & \sum_{p} \Bigg[\Gamma(\w_p) \label{eq:weak}\\
		 && + \alpha \sum_{I \in \weak} \ell(\bagL^p_I, \max_{i\in \R_I} \w_p^T \phi(\feat_i)) \Bigg]\nonumber
\end{eqnarray}
Note, we can decouple this optimization problem and independently solve for each category in our weakly labeled data set, $p \in \C_{\weak}$.
Let's consider a single category $p$. Our goal is to minimize the loss for category $p$ over images $I\in \weak$. 
We will do this by considering two cases. 
First, if $p$ is not in the weak label set of an image ($\bagL^p_I = -1$), then all regions in that image should be considered negative for category $p$. 
Second, if $\bagL^p_I = 1$, then we positively label a region $\feat_i$ if it has the highest confidence of containing object and negatively label all other regions. We perform the discovery of this top region in two steps. 
At first, we narrow down the set of candidate bounding boxes using the score, $\w_p^T \phi(\feat_i)$, from our fixed representation and detectors from the previous optimization step. 
This set is then refined to estimate the most region likely to contain the positive instance in a Latent SVM formulation. The implementation details are discussed section~\ref{sec:exp-minedboxes}.

Our final optimization step is to use the discovered annotations from our weak data-set to refine our detectors and feature representation from the previous optimization step. This amounts to the subsequent step for alternating minimization of the joint objective described in Equation~\ref{eq:joint}. We collectively utilize the strong annotations of images in $\strong$ and estimated annotations for weakly labelled set, $\weak$, to optimize for detector weights and feature representation, as follows: 
\begin{eqnarray}
\min_{ \substack{\w_k,\phi\\k \in \{\C, b\} }} &&\sum_{k} \Bigg[\Gamma(\w_k) \label{eq:final}\\
&&+ \alpha \sum_{I \in \{\weak\cup \strong\}} \sum_{i \in \R_I} \ell(\instL^k_i, \w_k^T\phi(\feat_i))\Bigg]\nonumber
\end{eqnarray}
This is achieved by re-finetuning the CNN architecture.

The refined detector weights and representation can be used to mine the bounding box annotations for weakly labeled data again, and this process can be iterated over (see Figure~\ref{fig:method}).
We discuss re-training strategies and evaluate the contribution of this final optimization step in Section~\ref{sec:exp-detection}.

\section{Experiments}
\label{sec:exp}
% experiments
We now study the effectiveness of our algorithm by applying it to a standard detection task.
\subsection{ILSVRC13 Detection Dataset \& Setup}
\label{sec:exp-dataset}
% Experiments dataset and setup

We use the ILSVRC13 detection dataset~\cite{ilsvrc} for our experiments. This dataset provides bounding box annotations for 200 categories. The dataset is separated into three pieces: train, val, test (see Table~\ref{table:dataset}). The training images have fewer objects per image on an average than validation set images, so they constitute classification style data~\cite{lsda}. Following prior work~\cite{rcnn}, we use the further separation of the validation set into val1 and val2. Overall, we use the train and val1 set for our training data source and evaluate our performance of the data in val2. 

\begin{table}
\centering
\begin{tabular}{c | c c}
\toprule
\multirow{2}{*}{Train} & Num images & 395905\\
& Num objects & 345854\\
\midrule
\multirow{2}{*}{Val} & Num images & 20121\\
& Num objects & 55502\\
\bottomrule
\end{tabular}
\caption{Statistics of the ILSVRC13 detection dataset. Training set has fewer objects per image than validation set.}
\label{table:dataset}
\end{table}

\begin{figure*}
\begin{center}

\begin{subfigure}[b]{.495\linewidth}
    \centering
    \includegraphics[width=\textwidth,height=2cm, natwidth=1165, natheight=229]{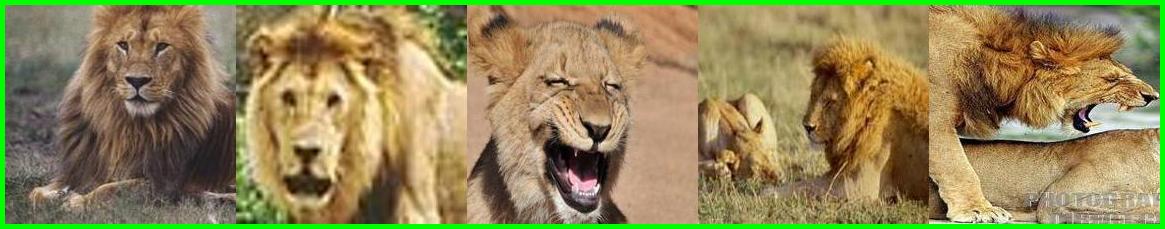}
\end{subfigure}\hfill
\begin{subfigure}[b]{.495\linewidth}
    \centering
    \includegraphics[width=\textwidth,height=2cm, natwidth=1265, natheight=196]{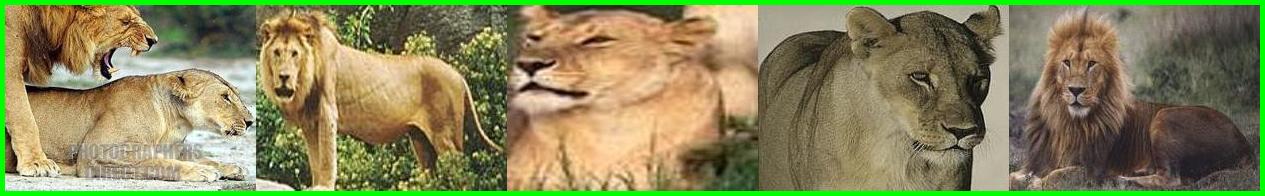}
\end{subfigure}\hfill

\begin{subfigure}[b]{.495\textwidth}
    \centering
    \includegraphics[width=\textwidth,height=2cm, natwidth=980, natheight=139]{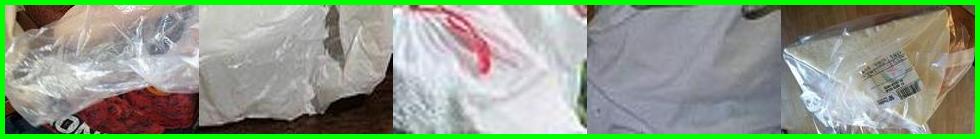}
\end{subfigure}\hfill
\begin{subfigure}[b]{.495\textwidth}
    \centering
    \includegraphics[width=\textwidth,height=2cm, natwidth=1395, natheight=267]{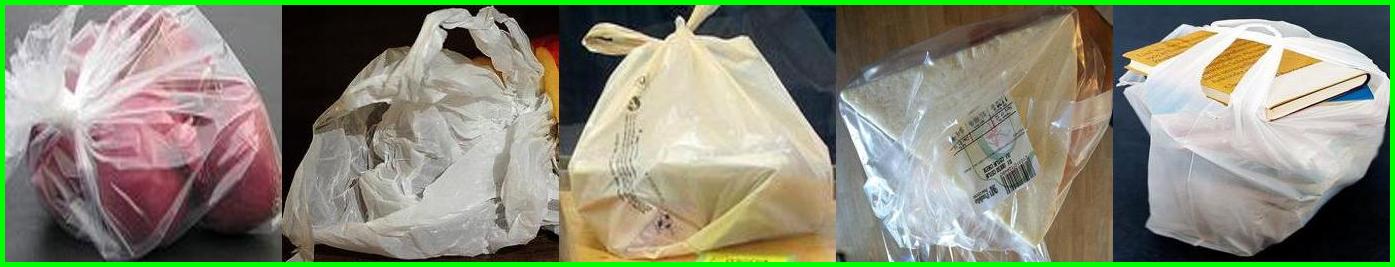}
\end{subfigure}\hfill

\begin{subfigure}[b]{.495\textwidth}
    \centering
    \includegraphics[width=\textwidth,height=2cm, natwidth=1010, natheight=163]{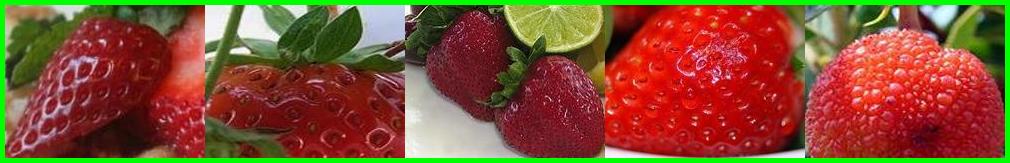}
\end{subfigure}\hfill
\begin{subfigure}[b]{.495\textwidth}
    \centering
    \includegraphics[width=\textwidth,height=2cm, natwidth=1135, natheight=223]{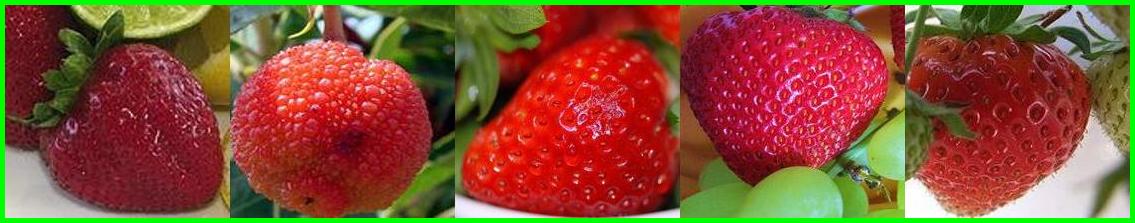}
\end{subfigure}\hfill

\begin{subfigure}[b]{.495\textwidth}
    \centering
    \includegraphics[width=\textwidth,height=3cm, natwidth=1280, natheight=305]{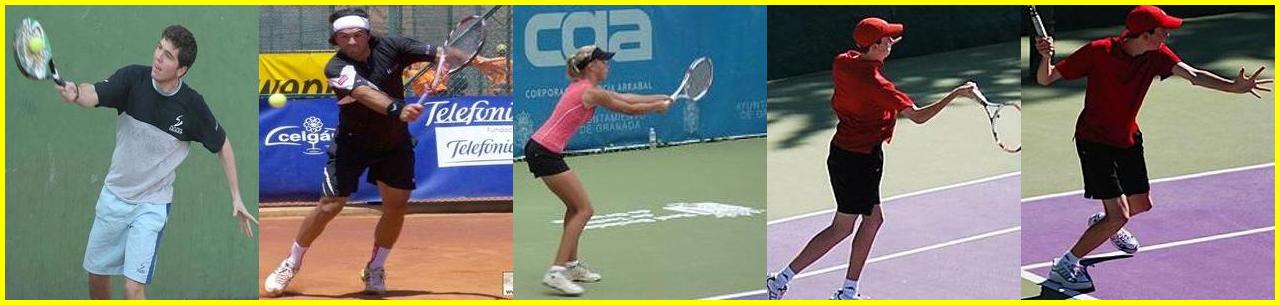}
\end{subfigure}\hfill
\begin{subfigure}[b]{.495\textwidth}
    \centering
    \includegraphics[width=\textwidth,height=3cm, natwidth=585, natheight=155]{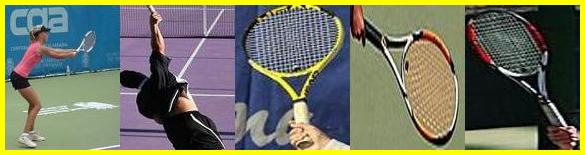}
\end{subfigure}\hfill
%\vspace{1mm}

\end{center}
   \caption{Example mined bounding boxes learned using our method. Left side shows the mined boxes after fine-tuning with images in classification settings only, and right side shows the mined boxes after fine-tuning with auxiliary strongly annotated dataset. We show top 5 mined boxes across the dataset for corresponding category. Examples with a  \textcolor{green}{green} outline are categories for which our algorithm was able to correctly mine patches of the object, while the feature space with only weak label training was not able to produce correct patches. In \textcolor{yellow}{yellow} we highlight the specific example of ``tennis racket". None of the discovered patches from the original feature space correctly located the tennis racket and instead included the person as well. After incorporating the strong annotations from auxiliary tasks, our method starts discovering tennis rackets, though still has some confusion with the person playing tennis. 
}
\label{fig:ex-mined-good}
\end{figure*}
%%%%%%%%%%%%%%%%%%%%%%%%%%%%%%%%%%%%%%%%%%%%%%%%%%%%%%%%%%%%%%%%%%%%%%%%%%%%%%%%%%%%%

Specifically, we use $\sim$1000 randomly chosen images per class from the train set for initializing our CNN weights. For this data we consider only have weak labels for all categories and train with the classification objective. We use the train set for this purpose as it tends to have more object-centric images and is therefore better suited to initializing classification weights. 

We have bounding box annotations for 100/200 of the categories in val1 ($\sim$5000 images with bounding boxes). Specifically, with the category names sorted alphabetically, categories 1-100 have strong annotations while 101-200 have only weak (image-level) annotations. Finally, we evaluate detection performance on the $\sim 10,000$ images in val2 across all 200 categories. 

We use open source deep learning framework, Caffe~\cite{caffe}, for the implementation, training and fine-tuning of our CNN architecture.

\subsection{Analysis of Discovered Positive Boxes}
\label{sec:exp-minedboxes}
% Analysis of mined boxes
One of the key components of our system is using strong annotations from auxiliary tasks to learn a representation where it's possible to discover patches that correspond to the objects of interest in our weakly labeled data source.
We begin our analysis by studying the patch discovery that our feature space enables. 
We optimize the patch discovery (Equation~\eqref{eq:weak}) using a one vs all Latent SVM formulation and optimize the formulation for AUC criterion~\cite{bilen2014_ijcv}.
The feature descriptor used is the output of the fully connected layer, $fc_7$, of the CNN which is produced after fine-tuning the feature representation with strongly annotated data from auxiliary tasks. 
Following our alternating minimization approach, these discovered top boxes are then used to re-estimate the weights and feature representations of our CNN architecture.

To evaluate the quality of mined boxes, we do precision analysis with respect to their overlap with ground truth which is measured using the standard intersection over union (IOU) metric. 
Table~\ref{table:pr-boxes} reports the precision for varying overlapping thresholds. 
Our optimization approach produces one positive patch per image with a weak label, and a discovered patch is considered a true positive if it overlaps sufficiently with the ground truth box that corresponds to that label. 
Since each patch, once discovered, is considered an equivalent positive (regardless of score) for the purpose of retraining, this simple precision metric is a good indication of the usefulness of our mined patches. 
It is interesting that a significant fraction of mined boxes have high overlap with the ground truth regions.  
For reference, we also computed the standard mean average precision over the discovered patches and report these results.

\begin{table*}
\centering
\begin{tabular}{l | c c c c | c}
\toprule
&\multicolumn{4}{c|}{Precision} & mAP\\
	& ov=0.3 & ov=0.5 & ov=0.7 & ov=0.9 & ov=0.5\\
	\midrule
Without auxiliary strong dataset  &  29.63 &  26.10 &  24.28 & 23.43 & 13.13\\
Ours & 32.69 & 28.81 & 26.27 & 24.78 & 22.81\\
\bottomrule
\end{tabular}
	\caption{Precision analysis and mAP performance of discovered patches in our weakly labeled training set (val1) of ILSVRC13 detection dataset. Comparison with varying amount of overlap with ground truth box. About 25\% of our mined boxes have an overlap of at least 0.9. Our method is able to significantly improve the quality of mined boxes after incorporating strong annotations from auxiliary tasks.}
	\label{table:pr-boxes}
\end{table*}

\begin{figure}
\begin{center}
\begin{subfigure}[b]{\linewidth}
    \centering
    \includegraphics[width=\linewidth, natwidth=1220, natheight=179]{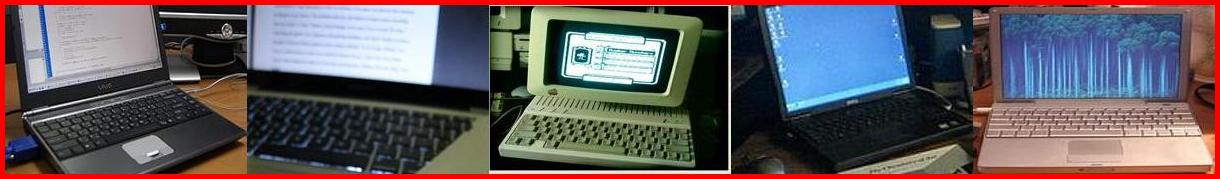}
\end{subfigure}\hfill
\begin{subfigure}[b]{\linewidth}
    \centering
    \includegraphics[width=\linewidth, natwidth=1220, natheight=179]{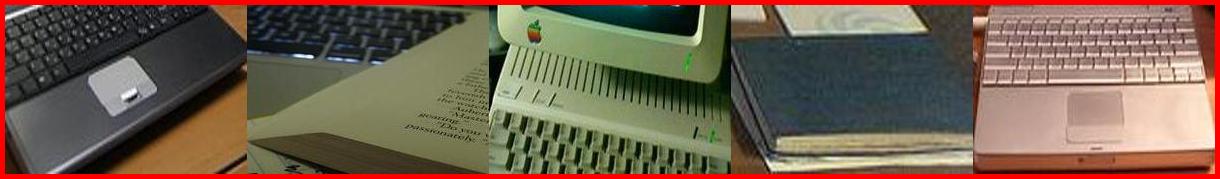}
\end{subfigure}\hfill
\end{center}
   \caption{Example mined boxes of the category ``laptop" where using auxiliary strongly annotated data causes patch discovery to diverge. \emph{Top row}: The mined boxes obtained after fine-tuning with images in classification settings only. \emph{Bottom row}: The mined boxes obtained after fine-tuning with the auxiliary strongly annotated dataset that contains the category ``computer keyboard". These patches were low scoring examples, but we show them here to demonstrate a potential failure case -- specifically, when one of the strongly annotated classes is a part of one of the weakly labeled classes. }
\label{fig:ex-mined-bad}
\end{figure}

It is important to understand not only that our new feature space improves the quality of the resulting patches, but also what type of errors our method reduces. In Figure~\ref{fig:ex-mined-good}, we show the top 5 scoring discovered patches before and after modifying the feature space with strong annotations from auxiliary tasks. We find that in many cases the improvement comes from better localization. For example without auxiliary strong annotations we mostly discover the face of a lion rather than the body that we discover after our algorithm. Interestingly, there is also an issue with co-occurring classes. In the bottom row of Figure~\ref{fig:ex-mined-good}, we show the top 5 discovered patches for ``tennis racket". Once we incorporate strong annotations from auxiliary tasks we begin to be able to distinguish the person playing tennis from the racket itself. Finally, there are some example mined patches where we reduce quality after incorporating the strong annotations from auxiliary tasks. For example, one of our strongly annotated categories is ``computer keyboard". Due to the strong training with keyboard images, some of our mined patches for ``laptop" start to have higher scores on the keyboard rather than the whole laptop (see Figure ~\ref{fig:ex-mined-bad}). 

\subsection{Detection Performance}
\label{sec:exp-detection}
\begin{figure}
\centering
\includegraphics[width=.5\linewidth]{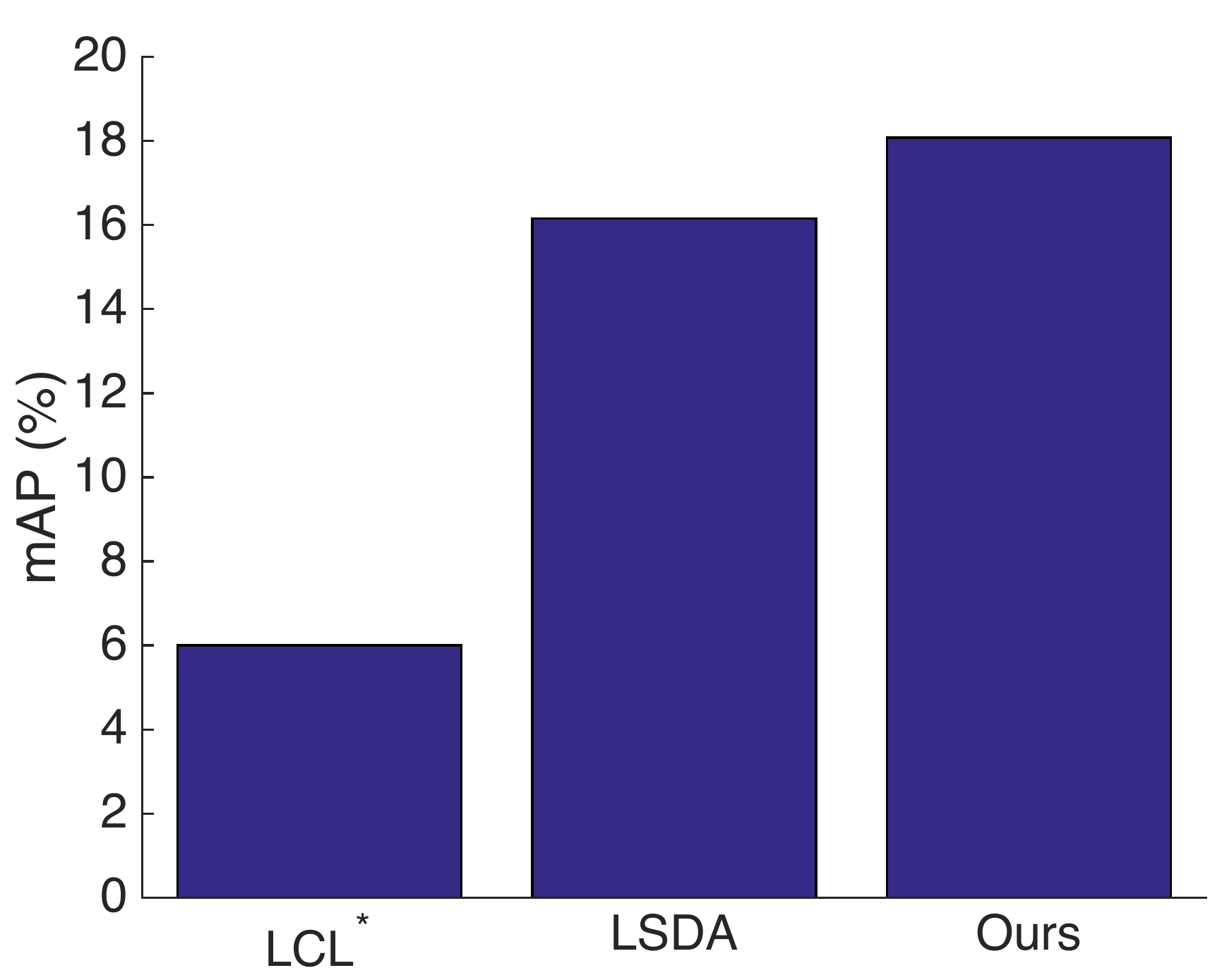}
\caption{Comparison of mAP (\%) for categories without any bounding box annotations (101-200 of val2) of ILSVRC13. Our method significantly outperforms both previous state-of-the-art algorithms: LCL~\cite{wang_eccv14} and LSDA~\cite{lsda}. *The value for LCL was computed across all 200 categories. Our experiments show this this is an easier task resulting in higher numbers overall.}
\label{fig:ap-all}
\end{figure}

% Analysis detection performance
Now that we've analyzed the intermediate result of our algorithm, we next study the full performance of our system. Figure~\ref{fig:ap-all} shows the mean average precision (mAP) percentage computed over the categories in val2 of ILSVRC13 for which we only have weakly annotated training data (categories 101-200). We compare to two state-of-the-art methods for this scenario and show that our algorithm significantly outperforms both of the previous state-of-the-art techniques. The first, LCL~\cite{wang_eccv14}, detects in the standard weakly supervised setting -- having no bounding box annotations for any of the 200 categories. This method also only reports results across all 200 categories. Our experiments indicate that the first 100 categories are easier on average then the second 100 categories, therefore the 6.0\% mAP may actually be an upper bound of the performance of this approach. The second algorithm we compare against is LSDA~\cite{lsda}, which does utilize the bounding box information from the first 100 categories. 

\begin{figure*}
\begin{center}
\centering
\includegraphics[width=\linewidth]{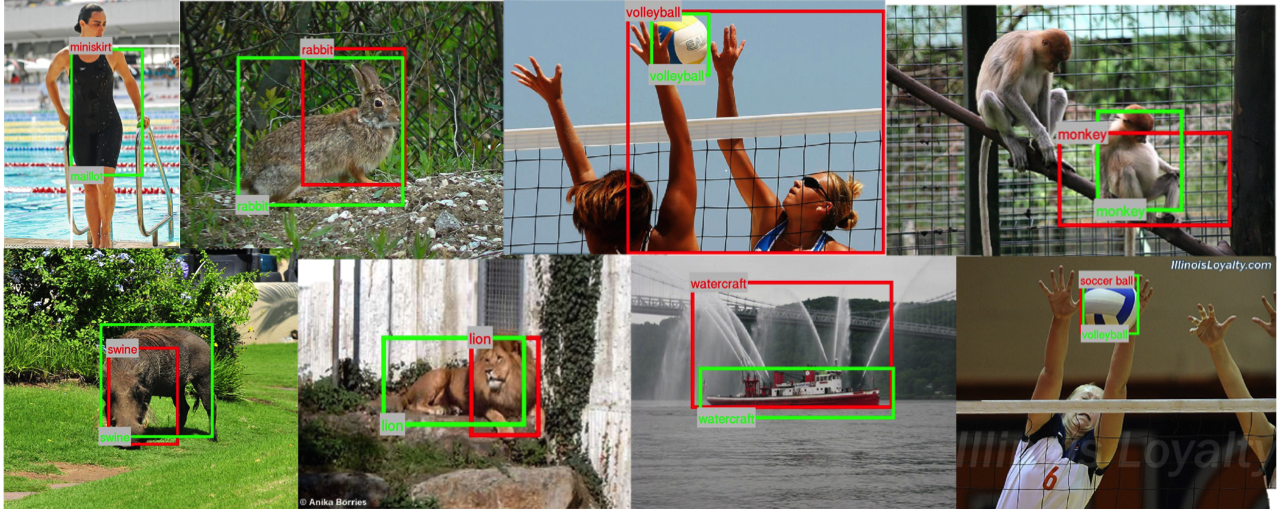}
\end{center}
   \caption{Examples where our algorithm outperforms the previous state-of-the-art. We show the top scoring detection from the baseline detector, LSDA~\cite{lsda}, with a \textcolor{red}{Red} box and label, and the top scoring detection from our method, LSDL, as a \textcolor{green}{Green} box and label. Our algorithm improves localization (ex: rabbit, lion etc), confusion with other categories (ex: miniskirt vs maillot), and confusion with co-occurring classes (ex: volleyball vs volleyball player)}
\label{fig:ex-wins}
\end{figure*}

We next consider different re-training strategies for learning new features and detection weights after discovering the positive patches in the weakly labeled data. Table~\ref{table:res-map} reports the mean average precision (mAP) percentage for no re-training (directly using the feature space learned after incorporating the strong labels), re-training only the category detection parameters, and retraining feature representations jointly with detection weights. 
In our experiments the improved performance is due to the first iteration of the overall algorithm.
We find that the best approach is to jointly learn to refine the feature representation and the detection weights. More specifically, we learn a new feature representation by fine-tuning all fully connected layers in the CNN architecture. 

\begin{table}
\centering
	\begin{tabular}{l c c c }
	\toprule
	& \multicolumn{3}{c}{Category Set}\\ \cline{2-4}
Re-train &	Weakly   & Strongly   &\multirow{2}{*}{All}  \\
Strategy &	 Labeled	& Labeled &  \\
\midrule
- & 15.85 & 27.81 & 21.83\\
detectors & 17.01 & 27.85 & 22.43 \\
rep+detectors & \bf 18.08 & 27.40 & 22.74\\
\bottomrule
	\end{tabular}
	\caption{Comparison of different ways to re-train after discovery of positive patches. We show mAP on val2 set from ILSVRC13. We find that the most effective way to re-train with discovered windows is to modify the detectors and the feature representation. }
	\label{table:res-map}
\end{table}

We finally analyze examples where our full algorithm outperforms the previous state-of-the-art, LSDA~\cite{lsda}. Figure~\ref{fig:ex-wins} shows a sample of the types of errors our algorithm improves on. These include localization errors, confusion with other categories, and interestingly, confusion with co-occurring categories. In particular, our algorithm provides improvement when searching for a small object (ball or helmet) in a sports scene. Training only with weak labels causes the previous state-of-the-art to confuse the player and the object, resulting in a detection that includes both. Our algorithm is able to localize only the small object and recognize that the player is a separate object of interest.

\section{Conclusion}
\label{sec:concl}
% Conclusion
We have presented a method which jointly trains a feature representation and detectors for categories with only weakly labeled data. We use the insight that strongly annotated detection data from auxiliary tasks can be used to train a feature representation that is conducive to discovering object patches in weakly labeled data. We demonstrate using a standard detection dataset (ImageNet-200 detection) that our method of incorporating the strongly annotated data from auxiliary tasks is very effective at improving the quality of the discovered patches. We then use all strong annotations along with our discovered object patches to further refine our feature representation and produce our final detectors. We show that our full detection algorithm significantly outperforms both the previous state-of-the-art methods which uses only weakly annotated data, as well as the algorithm which uses strongly annotated data from auxiliary tasks, but does not incorporate any MIL for the weak tasks. 

Upon acceptance of this paper, we will release all final weights and hyper parameters learned using our algorithm to improve the performance of the recently released >7.5K category detectors~\cite{lsda}.

{\small
\bibliographystyle{ieee}
\bibliography{2015-cvpr-lsda-mil}

\begin{thebibliography}{10}\itemsep=-1pt

\bibitem{alexe2010object}
B.~Alexe, T.~Deselaers, and V.~Ferrari.
\newblock What is an object?
\newblock In {\em Proc. CVPR}, 2010.

\bibitem{ali_cvpr14}
K.~Ali and K.~Saenko.
\newblock Confidence-rated multiple instance boosting for object detection.
\newblock In {\em IEEE Conference on Computer Vision and Pattern Recognition},
  2014.

\bibitem{andrews2002_nips}
S.~Andrews, I.~Tsochantaridis, and T.~Hofmann.
\newblock Support vector machines for multiple-instance learning.
\newblock In {\em Proc. NIPS}, pages 561--568, 2002.

\bibitem{pmt}
Y.~Aytar and A.~Zisserman.
\newblock Tabula rasa: Model transfer for object category detection.
\newblock In {\em IEEE International Conference on Computer Vision}, 2011.

\bibitem{bengio2009curriculum}
Y.~Bengio, J.~Louradour, R.~Collobert, and J.~Weston.
\newblock Curriculum learning.
\newblock In {\em In Proc. ICML}, 2009.

\bibitem{bilen2014_ijcv}
H.~Bilen, V.~P. Namboodiri, and L.~J. Van~Gool.
\newblock Object and action classification with latent window parameters.
\newblock {\em IJCV}, 106(3):237--251, 2014.

\bibitem{sentibank}
D.~Borth, R.~Ji, T.~Chen, T.~Breuel, and S.~F. Chang.
\newblock Large-scale visual sentiment ontology and detectors using adjective
  nown paiars.
\newblock In {\em ACM Multimedia Conference}, 2013.

\bibitem{verbeek2014_cvpr}
R.~G. Cinbis, J.~Verbeek, C.~Schmid, et~al.
\newblock Multi-fold mil training for weakly supervised object localization.
\newblock In {\em CVPR}, 2014.

\bibitem{daume}
H.~{Daum{\'e}~III}.
\newblock Frustratingly easy domain adaptation.
\newblock In {\em ACL}, 2007.

\bibitem{deselaers2012weakly}
T.~Deselaers, B.~Alexe, and V.~Ferrari.
\newblock Weakly supervised localization and learning with generic knowledge.
\newblock {\em IJCV}, 2012.

\bibitem{dietterich1997solving}
T.~G. Dietterich, R.~H. Lathrop, and T.~Lozano-P{\'e}rez.
\newblock Solving the multiple instance problem with axis-parallel rectangles.
\newblock {\em Artificial intelligence}, 1997.

\bibitem{decaf}
J.~{Donahue}, Y.~{Jia}, O.~{Vinyals}, J.~{Hoffman}, N.~{Zhang}, E.~{Tzeng}, and
  T.~{Darrell}.
\newblock {DeCAF: A Deep Convolutional Activation Feature for Generic Visual
  Recognition}.
\newblock In {\em Proc. ICML}, 2014.

\bibitem{duan-icml12}
L.~Duan, D.~Xu, and I.~W. Tsang.
\newblock Learning with augmented features for heterogeneous domain adaptation.
\newblock In {\em Proc. ICML}, 2012.

\bibitem{felzenszwalb2010_pami}
P.~F. Felzenszwalb, R.~B. Girshick, D.~McAllester, and D.~Ramanan.
\newblock Object detection with discriminatively trained part-based models.
\newblock {\em IEEE Tran. PAMI}, 32(9):1627--1645, 2010.

\bibitem{sa}
B.~Fernando, A.~Habrard, M.~Sebban, and T.~Tuytelaars.
\newblock Unsupervised visual domain adaptation using subspace alignment.
\newblock In {\em Proc. ICCV}, 2013.

\bibitem{rcnn}
R.~Girshick, J.~Donahue, T.~Darrell, and J.~Malik.
\newblock Rich feature hierarchies for accurate object detection and semantic
  segmentation.
\newblock In {\em In Proc. CVPR}, 2014.

\bibitem{gong-cvpr12}
B.~Gong, Y.~Shi, F.~Sha, and K.~Grauman.
\newblock Geodesic flow kernel for unsupervised domain adaptation.
\newblock In {\em Proc. CVPR}, 2012.

\bibitem{heckerman2013tractable}
D.~Heckerman.
\newblock A tractable inference algorithm for diagnosing multiple diseases.
\newblock {\em arXiv preprint arXiv:1304.1511}, 2013.

\bibitem{lsda}
J.~Hoffman, S.~Guadarrama, E.~Tzeng, R.~Hu, J.~Donahue, R.~Girshick,
  T.~Darrell, and K.~Saenko.
\newblock {LSDA}: Large scale detection through adaptation.
\newblock In {\em Neural Information Processing Systems (NIPS)}, 2014.

\bibitem{hoffman-iclr13}
J.~Hoffman, E.~Rodner, J.~Donahue, K.~Saenko, and T.~Darrell.
\newblock Efficient learning of domain-invariant image representations.
\newblock In {\em Proc. ICLR}, 2013.

\bibitem{caffe}
Y.~Jia, E.~Shelhamer, J.~Donahue, S.~Karayev, J.~Long, R.~Girshick,
  S.~Guadarrama, and T.~Darrell.
\newblock Caffe: Convolutional architecture for fast feature embedding.
\newblock {\em arXiv preprint arXiv:1408.5093}, 2014.

\bibitem{supervision}
A.~Krizhevsky, I.~Sutskever, and G.~E. Hinton.
\newblock Image{N}et classification with deep convolutional neural networks.
\newblock In {\em Proc. NIPS}, 2012.

\bibitem{kulis-cvpr11}
B.~Kulis, K.~Saenko, and T.~Darrell.
\newblock What you saw is not what you get: Domain adaptation using asymmetric
  kernel transforms.
\newblock In {\em Proc. CVPR}, 2011.

\bibitem{kumar2010self}
M.~P. Kumar, B.~Packer, and D.~Koller.
\newblock Self-paced learning for latent variable models.
\newblock In {\em In Proc. NIPS}, 2010.

\bibitem{lecun89}
Y.~LeCun, B.~Boser, J.~Denker, D.~Henderson, R.~Howard, W.~Hubbard, and
  L.~Jackel.
\newblock Backpropagation applied to handwritten zip code recognition.
\newblock {\em Neural Computation}, 1989.

\bibitem{pandey2011scene}
M.~Pandey and S.~Lazebnik.
\newblock Scene recognition and weakly supervised object localization with
  deformable part-based models.
\newblock In {\em Proc. ICCV}, 2011.

\bibitem{ilsvrc}
O.~Russakovsky, J.~Deng, H.~Su, J.~Krause, S.~Satheesh, S.~Ma, Z.~Huang,
  A.~Karpathy, A.~K. amd Michael~Bernstein, A.~C. Berg, and L.~Fei-Fe.
\newblock Imagenet large scale visual recognition challenge.
\newblock arXiv:1409.0575, 2014.

\bibitem{saenko-eccv10}
K.~Saenko, B.~Kulis, M.~Fritz, and T.~Darrell.
\newblock Adapting visual category models to new domains.
\newblock In {\em Proc. ECCV}, 2010.

\bibitem{overfeat}
P.~Sermanet, D.~Eigen, X.~Zhang, M.~Mathieu, R.~Fergus, and Y.~LeCun.
\newblock Overfeat: Integrated recognition, localization and detection using
  convolutional networks.
\newblock {\em CoRR}, abs/1312.6229, 2013.

\bibitem{singh2012unsupervised}
S.~Singh, A.~Gupta, and A.~A. Efros.
\newblock Unsupervised discovery of mid-level discriminative patches.
\newblock In {\em ECCV}. 2012.

\bibitem{siva2012defence}
P.~Siva, C.~Russell, and T.~Xiang.
\newblock In defence of negative mining for annotating weakly labelled data.
\newblock In {\em ECCV}. 2012.

\bibitem{siva2013looking}
P.~Siva, C.~Russell, T.~Xiang, and L.~Agapito.
\newblock Looking beyond the image: Unsupervised learning for object saliency
  and detection.
\newblock In {\em Proc. CVPR}, 2013.

\bibitem{song2014_icml}
H.~Song, R.~Girshick, S.~Jegelka, J.~Mairal, Z.~Harchaoui, and T.~Darrell.
\newblock On learning to localize objects with minimal supervision.
\newblock In {\em Proceedings of the International Conference on Machine
  Learning ({ICML})}, 2014.

\bibitem{song2014weakly}
H.~O. Song, Y.~J. Lee, S.~Jegelka, and T.~Darrell.
\newblock Weakly-supervised discovery of visual pattern configurations.
\newblock 2014.

\bibitem{wang_eccv14}
C.~Wang, W.~Ren, K.~Huang, and T.~Tan.
\newblock Weakly supervised object localization with latet category learning.
\newblock In {\em European Conference on Computer Vision (ECCV)}, 2014.

\bibitem{asvm}
J.~Yang, R.~Yan, and A.~G. Hauptmann.
\newblock Cross-domain video concept detection using adaptive svms.
\newblock {\em ACM Multimedia}, 2007.

\bibitem{thorsten2009_icml}
C.-N.~J. Yu and T.~Joachims.
\newblock Learning structural svms with latent variables.
\newblock In {\em Proc. ICML}, pages 1169--1176, 2009.

\bibitem{yuille2003_nc}
A.~L. Yuille and A.~Rangarajan.
\newblock The concave-convex procedure.
\newblock {\em Neural Computation}, 15(4):915--936, 2003.

\bibitem{zeiler-arxiv-2013}
M.~{Zeiler} and R.~{Fergus}.
\newblock {Visualizing and Understanding Convolutional Networks}.
\newblock {\em ArXiv e-prints}, 2013.

\bibitem{zhang2005multiple}
C.~Zhang, J.~C. Platt, and P.~A. Viola.
\newblock Multiple instance boosting for object detection.
\newblock In {\em Advances in neural information processing systems}, 2005.

\end{thebibliography}
}

\end{document}